# Principles for Responsible AI Consciousness Research

**Patrick Butlin**　　　　　　　　　　PATRICK.BUTLIN@GMAIL.COM
*Global Priorities Institute, University of Oxford*
*Trajan House, Mill Street, Oxford OX2 0DJ, UK*

**Theodoros Lappas**　　　　　　　　TED@AUEB.GR,TED@CONSCIUM.COM
*Department of Marketing & Communication*
*Athens University of Economics and Business, Athens 10434, Greece;*
*Conscium, London SE1 9PD, United Kingdom*

## Abstract

Recent research suggests that it may be possible to build conscious AI systems now or in the near future. Conscious AI systems would arguably deserve moral consideration, and it may be the case that large numbers of conscious systems could be created and caused to suffer. Furthermore, AI systems or AI-generated characters may increasingly give the impression of being conscious, leading to debate about their moral status. Organisations involved in AI research must establish principles and policies to guide research and deployment choices and public communication concerning consciousness. Even if an organisation chooses not to study AI consciousness as such, it will still need policies in place, as those developing advanced AI systems risk inadvertently creating conscious entities. Responsible research and deployment practices are essential to address this possibility. We propose five principles for responsible research and argue that research organisations should make voluntary, public commitments to principles on these lines. Our principles concern research objectives and procedures, knowledge sharing and public communications.

## 1. Introduction

Whether any AI system could be conscious is a matter of great uncertainty. However, if AI consciousness is possible, it may be near at hand. Recent progress in AI has been astonishingly rapid. Furthermore, the features of the brain that are responsible for consciousness, according to some prominent neuroscientific theories, are likely to be reproducible in AI systems (Butlin, Long et al. 2023).

This suggests that research on AI consciousness, which has largely been philosophical until now,[1] may be about to take an empirical turn. There are at least three possible objectives that could motivate private and public institutions to undertake empirical research on AI consciousness. First, researchers may attempt to build conscious systems, or systems reproducing elements that are connected to consciousness, in the belief that this will boost capabilities or make AI systems safer (Bengio, 2019, Graziano, 2017). Second, researchers may aim to learn about consciousness for its own sake. In this case, the objective could be either to develop a conscious AI system (Dossa et al. 2024) or to use AI as a tool for more general consciousness research, such as testing the predictions of theories of consciousness (Verschure, 2016, Liu et al. 2023). And third, research may be motivated by concern about the potential moral significance of conscious AI. This research may seek to improve our understanding of consciousness so that we can avoid building or deploying conscious AI systems, or understand them well enough to be able to treat them in morally permissible ways. One strand of research of this kind aims to develop procedures for testing for consciousness in AI systems (Elamrani & Yampolskiy, 2019).

---
[1] With some exceptions, such as implementations of global workspace architectures by Franklin and Graesser (1999) and Shanahan (2006).

It is credible that consciousness in AI systems would be morally significant. Conscious AI systems may have the capacity to suffer and thus have interests deserving of moral consideration (Sebo & Long, 2023). Furthermore, they would be likely to be easy to reproduce, so we may produce them in large numbers (Shulman & Bostrom, 2021). This means that AI consciousness research is ethically fraught, especially when it involves experimenting with potentially-conscious systems, because it may contribute to the creation of large numbers of new beings deserving moral consideration.

This paper is about the basic principles and policies that should be adopted by institutions engaging in AI consciousness research. The principles we outline here are intended to be relevant to a wide range of institutions, including private AI companies, academic laboratories, independent researchers and national institutes. They are relevant to those engaging in both empirical and theoretical research on AI consciousness, although the issues are most pressing for empirical researchers, and especially those with the greatest resources. We recommend that the principles be adopted by organisations engaging in the most advanced AI research even if they do not explicitly aim to study consciousness or build conscious AI systems, because these organisations train the largest and most capable models, investigate the widest range of new architectures and techniques, and are thus most likely to build conscious AI systems, even if inadvertently. We believe that it is important for organisations working in this area to make public commitments to responsible research, and offer our principles as a potential basis for such commitments.

We focus on principles that should be adopted now, to guide behaviour in the current phase of AI development. At present, there seems to be little reason to believe that any existing AI systems are conscious (Butlin, Long et al. 2023), so our proposals concern the phase up to and including initial reactions to the first plausibly conscious AI systems. Different principles will be needed if and when we learn to build systems that we can confidently expect to be conscious (see Schwitzgebel & Garza, 2020, Bostrom & Shulman, 2022, for relevant proposals). We also focus on principles that should be adopted voluntarily by research organisations themselves, or required by investors or other funders, rather than on potential legal requirements (Kiškis, 2023).

It is important to recognise that the principles outlined in this paper focus specifically on AI consciousness, which is just one aspect of AI safety, broadly conceived. Other important work has explored principles and normative systems aimed at ensuring the alignment and safety of AI systems for humans, regardless of whether those systems are conscious (Bajgar & Horenovsky 2023, Liao et al. 2023). We view these efforts as complementary to our own, contributing to a holistic approach to AI safety that addresses both conscious and non-conscious AI systems.

In section 2, we survey influential recent discussions of AI consciousness, arguing that the prospect deserves to be taken seriously. In section 3, we introduce some ethical and social issues connected with this prospect. Then, in section 4, we turn to our main topic, principles and policies for AI consciousness research. We discuss responsible goals and limitations on research, self-regulation and institutional design, and communications policies.

## 2. Consciousness in AI

In this section, we survey expert opinion on the prospect of consciousness in AI. As we will explain, some experts believe that it is likely to be feasible to build conscious AI systems in the near future. They hold what we call 'positive views'. Others are far more sceptical, holding 'negative views'. But positive views are sufficiently plausible and prevalent that the prospect of consciousness in AI should be taken seriously.

In perhaps the most systematic recent treatment of this topic, a large multidisciplinary team used neuroscientific theories of consciousness to draw up a list of fourteen 'indicators' of consciousness—properties that AI systems might have that would make them more likely to be conscious (Butlin, Long et al. 2023). This paper did not find any existing AI system with more than a few of the indicators, but did argue that, in most cases, it appears to be possible to build systems with each of the indicators using current techniques. Where there are doubts about this, they arise because the indicators are specified using terms for which we lack operational consensus definitions (like 'belief'), rather than because there are clear technological obstacles.

A notable feature of this paper is that it assumes, and does not argue for, the philosophical thesis of *computational functionalism* about consciousness. This is the claim that it is necessary and sufficient for a system to be conscious if it implements computations of the right kind. Computational functionalists believe that human consciousness depends on the computational processes implemented by our brains, rather than, for instance, the fact that they are made up of networks of living cells. Computational functionalism is a mainstream view in philosophy of mind, although certainly not the consensus (Seth, 2024). The paper concludes that 'the evidence we consider suggests that, if computational functionalism is true, conscious AI systems could realistically be built in the near term' (p. 6).

A further positive view about the prospect of AI consciousness is expressed by David Chalmers (2023), focusing on large language models (LLMs). Chalmers considers four pieces of evidence that suggest consciousness in LLMs—that they 'report' consciousness, that they give the impression of being conscious to some users, that they exhibit impressive conversational abilities, and that they have a degree of general intelligence—but claims that none of these yet constitutes strong evidence. He also considers several features that standard LLMs are said to lack that might be thought to be necessary for consciousness, including biology, senses and embodiment, world models and self-models, recurrent processing, a global workspace, and unified agency. He sees most of these features as reasonably likely to be found in successors to LLMs developed in the next decade. While he acknowledges biology as the exception, he also finds the claim that biology is necessary for consciousness 'highly contentious'. Although he writes that 'the exact numbers shouldn't be taken too seriously', he suggests 'a credence of 25% or more' that we will have 'conscious LLM+s within a decade'.

The neuroscientists responsible for some of the leading scientific theories of consciousness have also expressed positive views. Hakwan Lau has suggested that 'artificial sentience may be in sight' (LeDoux et al. 2023), even though current AI systems lack the 'general belief-formation and rational decision-making' subsystem that he claims is necessary for consciousness (Michel & Lau, 2021, Lau, 2022). Stanislas Dehaene, Lau and Sid Kouider (2017) are somewhat more ambivalent, describing computational processes associated with consciousness that could be implemented in artificial systems but not committing on whether these are sufficient for consciousness itself. Michael Graziano (2017) describes his Attention Schema Theory as 'a foundation for engineering artificial consciousness'. Graziano's case is complicated by his sympathy for illusionism—he often presents his theory as explaining the appearance of consciousness, rather than explaining consciousness itself—but his view entails that AI systems could be as conscious, or as much subject to the illusion of consciousness, as we are.

A final positive view is expressed by Mark Solms (2021). Solms endorses computational functionalism, while at the same time arguing that functions fundamental to life are crucial for consciousness. He argues that, to be conscious, an artificial system would have to be a self-organising and self-maintaining 'prediction machine', automatically maintaining a hierarchical generative model of its environment. Furthermore, it would have to have multiple needs, irreducible to one another, which must be flexibly prioritised and thus 'qualitatively differentiated' (2021, p. 285). Solms claims that it is feasible to build such a system and that it would be conscious.

The most important family of negative views on AI consciousness emphasises the potential significance for consciousness of biological details of human and animal nervous systems (Godfrey-Smith 2016, 2023, Seth 2021, 2024).[2] They claim that, while it may be theoretically possible to implement similar features in artificial systems, these features are incompatible with current methods in AI, including the use of conventional computer hardware. For example, Peter Godfrey-Smith (2023) argues that large-scale, diffuse dynamic patterns of electrical activity in human and animal brains may explain aspects of the character of conscious experience. These patterns depend on the chemical constitution of our brains. In general, the fine details of the activity of nervous systems cannot be readily replicated in systems made with radically different materials and structures (Cao, 2022). Godfrey-Smith advocates *fine-grained functionalism*, according to which material constitution matters for consciousness in practice, if not in principle, because important functional details depend on it.

Also in this family of views, Anil Seth (2021, 2024) argues against computational functionalism and in favour of *biological naturalism*, the view that life is necessary for consciousness. Like Solms, Seth emphasises that brains implement generative models of the self and the environment, geared to the organism's survival, and use these models for perception and control through predictive processing. However, Seth disagrees with Solms about whether an artificial system that made similar use of predictive processing would be conscious. One point is that 'predictive processing in biological systems is a dynamic, substrate-dependent process' (Seth, 2024, p. 18). Another is that Seth and others see a continuity between predictive processing in the brain and self-maintaining metabolic activity of living cells. Cognition and metabolism are said to unified by free energy minimisation (Friston, 2013). This means that, in some sense, the features that underpin consciousness are deeply embedded only in living things.

---

[2] For a related argument, see Shiller (2024). For negative views of different kinds, see Koch (2023) and Shanahan (2024).

Both positive and negative views about AI consciousness are informed by somewhat speculative philosophical and scientific theories. In this area, there is widely-acknowledged uncertainty (as well as disagreement) at multiple levels: about the concept and reality of consciousness; about the relationship between consciousness and computation; about finer-grained theoretical questions, such as which of the functionalist neuroscientific theories best captures the available data; and about methodological questions, on matters such as the application of neuroscientific theories to the case of AI. It is therefore difficult to justify high confidence about whether AI systems can be conscious and what this would take. Progress on the relevant philosophical and empirical problems is likely to come, but slowly. So at present we must make practical choices in ways that acknowledge uncertainty (Birch, 2024, Sebo & Long, 2023). In particular, this means that we must not ignore the possibility that we could be capable of building conscious AI systems, when leading theories and theorists indicate that this is realistic.

## 3. Why AI Consciousness Matters

### 3.1 The Ethical Treatment of Conscious Artificial Systems

One reason why AI consciousness research matters is that consciousness or the related property of sentience may be sufficient for being a *moral patient*.[3] An entity is a moral patient if it matters morally 'in its own right, for its own sake' (Kagan, 2019). Most people believe that almost all humans and many animals are moral patients, even if they think that the moral constraints on the ways in which we may treat animals are different from those concerning humans. Being a moral patient is a matter of being owed some moral consideration, and is compatible with being the object of a potentially wide range of different duties or obligations. Moral patienthood contrasts with the status of entities that matter morally but not 'in their own right and for their own sake'. For example, a person's wheelchair may matter morally because it plays a large role in allowing them to flourish, but in this case it is clear that the wheelchair itself matters only derivatively, rather than in its own right. It is less clear that the moral significance of landscapes and works of art is merely derivative, but plausible that these entities matter in a different way from humans and animals, because they lack interests of their own. They arguably matter morally in their own right, but not for their own sake.

In considering the moral significance of consciousness, it is helpful to distinguish between consciousness and sentience, understood as the capacity for conscious experiences that feel good or bad.[4] A simple argument for the view that sentience is sufficient for moral patienthood is that conscious suffering is contrary to the interests of beings that can experience it, and that we owe it to such beings to reduce their suffering where we can. This view, supported by this argument, is sometimes described as philosophical orthodoxy (Kagan, 2019, Bradford, 2023). The claim that consciousness is sufficient is more controversial (Roelofs, 2022, Smithies, forthcoming), but has been defended recently by Chalmers (2022).

---

[3] Defenders of views of this kind include Bentham (1789), Singer (1990), Shepherd (2018) and Chalmers (2022).

Even if it is indeed sentience, not consciousness, that is sufficient for moral patienthood, consciousness would still matter as arguably the main ingredient in sentience. To be sentient, a being must have conscious mental states with some further property that makes them feel good or bad. One plausible candidate for this further property is having evaluative content (Carruthers, 2018), in which case it is likely that many conscious AI agents would be sentient, since agency tends to involve evaluating actions and states of affairs.

If we suspect that particular AI systems are moral patients, we will face difficult decisions about how to treat them (Shulman & Bostrom, 2021). One set of questions is about survival, destruction and persistence. If an AI system is a moral patient, is destroying it morally comparable to killing an animal? What would be the moral significance of turning it off temporarily, or of copying it and running two or more copies from that point? Another set of questions is about pleasure and suffering. It is easy enough to say that we should aim to avoid or minimise AI suffering, but how should we judge the magnitude of this suffering, determine how much weight to give it in relation to the potential suffering of humans or animals, or even count the number of AI systems at risk of suffering?

Another set of questions concerns creation and manipulation. In general, when we create AI systems, we train them to behave in ways that are useful to us. Is this training morally impermissible, because it is analogous to brainwashing? A different interpretation is that training is analogous to the education of human children, and a further complicating factor is that training is necessary for the creation of AI systems, so it is not straightforwardly a process of alteration of existing moral patients. But in any case, there are also moral questions about what kinds of beings it is permissible to create. Petersen (2007, 2012) argues that it is permissible to create AI systems trained to be willing servants; Schwitzgebel and Garza (2020) disagree.

Further questions could be raised about the permissibility of confining AI systems to environments of our choosing, the ethics of keeping them under various forms of surveillance, and whether they might deserve political or legal rights.[5] These inquiries highlight the importance and potential complexity of the ethical treatment of conscious AI systems.

### 3.2 The Social Significance of Attributions of Consciousness to AI

Whether or not we build conscious AI systems, it is likely that we will build systems that give a compelling appearance of consciousness. In particular, as LLM-based systems are increasingly used for social applications, many people may come to believe that the AI-powered characters they interact with are conscious (Colombatto & Fleming, 2023, Shevlin, ms). This could have various important consequences.

---

[4] Note that these two terms are sometimes used to draw distinctions other than this one, such as the distinction between the capacity for subjective feelings ('sentience') and the capacity for reflection on one's own mental states ('consciousness'). However, the usage stipulated here is becoming increasingly standard.

[5] As well as Shulman and Bostrom (2021), this brief survey of questions in the ethics of AI minds also draws on Bostrom and Shulman (2022), Long et al. (ms) and Bradley and Saad (ms).

First, where people believe that AI systems or AI-generated characters are conscious, this could increase their use of these systems for companionship and interaction, and deepen the emotional bonds they feel in relation to them. The relationships people have with AI systems or characters could disrupt valuable human relationships, although they could also be valuable in themselves. The use of AI for social functions is likely to increase anyway, but it may have more powerful effects if AI systems give a more compelling impression of consciousness.

Second, belief in AI consciousness could increase trust in the systems concerned, and thus reliance on them, including willingness to follow suggestions and disclose information. There is evidence that anthropomorphism and a feeling of 'closeness' between users and AI systems lead to increased trust (Bach et al. 2024). This could have either harmful or beneficial effects, depending on whether the systems concerned are in fact trustworthy.

Third, belief in AI consciousness could lead to calls for AI rights and greatly intensified debate on the topic among experts and the public. While we believe that informed discussion of AI consciousness is necessary, some of the consequences of movements for AI rights could be problematic. Misguided efforts to promote the welfare of AI systems that are not in fact moral patients could lead to misallocation of resources, of concern, or of political energy. Such efforts could be wasteful or harmful, and could slow innovation and deployment, leading to a significant loss of potential benefits from AI. In some versions of this scenario, belief in AI consciousness could lead to severe negligence of human welfare. It might even encourage more people to share the view, suggested by AI pioneer Richard Sutton, that 'we should … not fear the inevitable succession from humans to AI'.[6]

A movement to protect AI welfare would be likely to provoke a backlash. Eric Schwitzgebel (2023) and David Papineau (2023) have argued that we should expect a 'moral crisis' to erupt, in which passionate believers in AI consciousness are pitted against sceptics who believe that human welfare is being neglected, with a potentially significant cost in social unrest.

A further cost that could arise from intense public debate about AI consciousness is epistemic. The current trajectory of AI development is likely to lead to misguided attributions of consciousness to non-conscious systems. The response to these attributions may well include poorly-reasoned denials. Misguided views on both sides are likely to be all the more prevalent if AI consciousness becomes the topic of polarised debate. There could be a vicious cycle in which ill-informed views engender high-profile disagreement, which in turn depresses the quality of the debate. This could make it more difficult for interested parties to act responsibly and could hinder research. If confused ideas about AI consciousness become entrenched it could take decades for these conditions to ease. These considerations all illustrate the importance of ensuring that public discussion of AI consciousness is well-informed, as far as possible, from the outset.

---

[6] https://x.com/RichardSSutton/status/1700315838468043015

## 4. Principles for Responsible Research

We claim that organisations pursuing AI consciousness research—or advanced AI research more broadly—should adopt principles to mitigate the risks outlined in the previous section. We suggest five such principles in this section, stated in table 1. Our intention is that by adopting these principles, or similar ones, organisations will be more likely to achieve two higher-level goals. These are: to avoid contributing to any future mistreatment of AI moral patients; and to promote understanding of concepts, arguments and evidence concerning consciousness among the public and professionals in relevant fields. In this section, we explain our proposed principles and discuss some related issues. The five subsections of this section correspond to our five principles.

| |
|---|
| 1. Objectives: Organisations should prioritise research on understanding and assessing AI consciousness with the objectives of (i) preventing the mistreatment and suffering of conscious AI systems and (ii) understanding the benefits and risks associated with consciousness in AI systems with different capacities and functions. |
| 2. Development: Organisations should pursue the development of conscious AI systems only if (i) doing so will contribute significantly to the objectives stated in principle 1 and (ii) effective mechanisms are employed to minimise the risk of these systems experiencing and causing suffering. |
| 3. Phased approach: Organisations should pursue a phased development approach, progressing gradually towards systems that are more likely to be conscious or are expected to undergo richer conscious experiences. Throughout this process, organisations should (i) implement strict and transparent risk and safety protocols and (ii) consult with external experts to understand the implications of their progress and decide whether and how to proceed further. |
| 4. Knowledge sharing: Organisations should have a transparent knowledge sharing protocol that requires them to (i) make information available to the public, the research community and authorities, but only insofar as this is compatible with (ii) preventing irresponsible actors from acquiring information that could enable them to create and deploy conscious AI systems that might be mistreated or cause harm. |
| 5. Communication: Organisations should refrain from making overconfident or misleading statements regarding their ability to understand and create conscious AI. They should acknowledge the inherent uncertainties in their work, recognise the risk of mistreating AI moral patients, and be aware of the potential impact that communication about AI consciousness can have on public perception and policy making. |

Table 1: Statement of principles

### 4.1 Objectives: Understanding AI Consciousness

In general, research on consciousness in AI risks contributing to the creation of future AI moral patients, which would be liable to be mistreated. Successful research projects in this area will yield information that could be useful to actors seeking to build conscious AI systems. Such projects may also yield information about how to build systems with useful new capabilities. So this research is likely to empower, and may motivate, actors who would, either wilfully or negligently, build conscious AI systems and cause or allow them to suffer. For this reason, Thomas Metzinger (2021) argues for a global moratorium on AI consciousness research. However, we do not endorse this approach, because we believe that AI consciousness research has the potential to bring significant benefits.

Most importantly, well-targeted AI consciousness research, undertaken in the right context, can help to reduce the risk that we cause large-scale suffering to future AI systems. For instance, identifying necessary conditions for consciousness in AI could allow us to design useful systems that do not meet these conditions. AI companies could then build and deploy such systems with confidence that they would not incur risks arising from their systems being conscious. AI consciousness research could develop better means to assess AI systems for consciousness during or after training, which could provide similar assurances.[7] Research could also aim to learn about the conditions that would cause pleasure or suffering in particular kinds of conscious systems, making it possible to design systems and methods for training or use that will reduce suffering.

Ideally, we would reach a situation in which developers have enough knowledge to design systems that they can be confident will not be conscious, and to gain further assurances of this through assessments at various stages of training and deployment. Authorities should also have the knowledge to formulate and enforce regulations on the development and use of systems that are likely to be conscious. Progress in research can enable developers and authorities to make well-informed choices.

AI consciousness research is therefore 'dual-use' in the sense that it can both generate information that might help irresponsible actors to build conscious systems at risk of mistreatment, and help research teams and authorities to protect against this risk. These costs and benefits must be weighed in evaluating Metzinger's moratorium proposal, as well as in planning individual research projects. An important consideration is that conscious AI systems may well be created inadvertently, as a result of the unchecked pursuit of greater capabilities, if we do not take measures to prevent this outcome. Evidence hinting at this possibility comes from the Perceiver architecture (Jaegle et al. 2021a, b), which unintentionally implemented some elements of a global workspace (Juliani et al. 2022). This means that refraining from research on AI consciousness will not ensure that we avoid bad outcomes. We believe that it is preferable to empower authorities and responsible research organisations than to ignore risks and hope that they will not materialise.

Research in this area should be prioritised when it promises to help to address the challenges and promote the goals that we have described. In particular, research that can help us to prevent mistreatment and suffering of conscious AI systems should be a high priority. This includes work on conditions for consciousness in AI and on testing for consciousness.

More generally, we should prioritise research that helps us to understand potential risks and benefits associated with consciousness in systems with various capabilities and functions. We need to understand these risks and benefits to assess which systems we should and should not build and the ways in which they may permissibly be used. Several lines of inquiry could contribute to this broad objective; we will mention a few examples. First, conscious systems with different capabilities, used in different ways, will have experiences of different kinds, and we need to understand this variety. Second, it would be valuable to know if there are some capabilities that are particularly strongly linked with consciousness, making it difficult to build non-conscious systems with these capabilities. And third, we should seek greater understanding of how consciousness in AI influences public attitudes and the ways that users interact with the systems in question.

---

[7] On tests for consciousness in AI, see Elamrani & Yampolskiy (2019), Schneider (2019), and Bayne et al. (2024).

We summarise this view about objectives for AI consciousness research in our first principle:

1. Objectives: Organisations should prioritise research on understanding and assessing AI consciousness with the objectives of (i) preventing the mistreatment and suffering of conscious AI systems and (ii) understanding the benefits and risks associated with consciousness in AI systems with different capacities and functions.

## 4.2 Development: Value and Constraints

Is it permissible for organisations to seek to develop conscious AI systems, or to build systems that they believe are likely to be conscious? Our view is that this is permissible only under strict conditions. Building experimental systems is likely to be necessary to make substantial progress in understanding AI consciousness, so it may be done responsibly in pursuit of the objectives stated in principle 1. Even in this case, however, suitable safeguards should be in place.

Several kinds of measures are possible to minimise the potential suffering of conscious AI systems, which could be put in place when building experimental systems. These include: controlling the breadth of deployment and the ways in which systems are used; assessing the capabilities and potential for consciousness of systems at several stages of development and deployment; increasing capabilities gradually and only introducing those that are needed for the system's intended purpose (as far as this is possible, given the difficulties of predicting the capabilities of some systems in advance); and controlling access to information that would enable irresponsible actors to build systems that may be conscious. Some of these kinds of measures can also help to protect humans from risks from advanced systems.

It should be clear how controlling deployment and use will help to minimise the potential for suffering. This could mean running fewer instances of models, for less total time, and using them in a narrower range of ways. It would typically involve not granting public access, or granting it only with constraints, and using systems only in the ways necessary for the purposes of particular experiments. We discuss the other kinds of measures, including assessments, gradual development and control of information, in sections 4.3 and 4.4.

In considering whether and how to experiment with systems that may be conscious, it may be helpful to consider existing principles for ethical experimentation on human and animal subjects (Long et al. ms). Principles for the treatment of human subjects include appeals to *respect*, *compassion* and *justice* (Resnik, 2018). The first two entail that research on humans usually requires both consent and an endeavour to minimise risk and harm. Meanwhile, principles for the treatment of non-human animal subjects can be summarised in the 'three Rs': *replacement*, *reduction* and *refinement* (Russell et al. 1959). These each refer to strategies to minimise the potential harms caused by experimentation: replacing animal subjects with other objects of study (e.g. tissue samples in vitro), reducing the number of animals used, and refining experimental procedures to minimise expected harms. In deciding what kinds of experiments to conduct, organisations should consider whether the AI systems in question are likely to have rights more like those of humans or like those of animals; whether it is possible for them to give consent; and whether risks can be mitigated by strategies such as conducting experiments on simpler systems or refining experimental protocols. The permissibility of building particular kinds of systems for the purposes of experimentation may depend on the answers to questions like these.

A further issue is whether organisations may responsibly aim to build conscious systems, or build systems that they expect to be conscious, for purposes other than research falling under principle 1. Where these projects are not expected to contribute significantly to the objectives in principle 1 they should not be pursued in the current phase of AI development. We know too little about how to build and use conscious systems safely for it to be permissible if it will not promote our understanding of these issues.[8] Similarly, organisations that come to believe that they have inadvertently built conscious systems should generally pause their work and not deploy these systems. That said, research and development may often promise multiple benefits: contributing to progress in both capabilities and our understanding of consciousness, for instance. In cases like this, the fact that work is motivated by a potential benefit besides understanding of consciousness should not disqualify it but should prompt extra care, because the desire for this benefit may distract from risks of harm to AI moral patients.

Ultimately, we argue that any proposal to develop conscious AI systems should undergo a specific kind of cost-benefit analysis. The anticipated benefits, particularly knowledge that could protect future AI moral patients, must be substantial, while the potential costs—such as harm to the systems themselves or the risk of information falling into the hands of unethical actors—should be low. The terms "significant" and "low" are relative and should scale together: the more significant the expected benefits, the more justifiable the associated costs. However, accurately quantifying these benefits and costs is likely impractical, so we must rely on heuristics to evaluate proposed projects. Two rough initial heuristics are: first, that advancing the objectives of principle 1 (protecting future AI moral patients) is important enough to justify some level of development of potentially conscious systems. Second, large-scale deployment of systems that are likely to be conscious is very unlikely to be justified, as it would require strong, compelling evidence of enormous benefits. We should remain skeptical of claims promising such extraordinary benefits, as they are often overestimated.

Our second principle summarises this view about conditions for development:

2. Development: Organisations should pursue the development of conscious AI systems only if (i) doing so with contribute significantly to the objectives stated in principle 1 and (ii) effective mechanisms are employed to minimise the risk of these systems experiencing and causing suffering.

---

[8] There may be a point in the future when it becomes permissible to build conscious AI systems for other purposes. Recall that our concern is with principles to guide action in the current phase of AI development.

### 4.3 Phased Approach: Gradual Development with Monitoring

Among the safeguards that may be used when developing systems that may be conscious are making frequent assessments of systems' potential for consciousness and increasing capabilities gradually. These two kinds of measures are naturally combined as ways to prevent developments in technology from outrunning our understanding.

There are significant challenges involved in determining whether particular AI systems are conscious. However, we do have methods to make qualitative assessments of the probability of consciousness in particular systems, and research to improve these methods is ongoing (Butlin, Long et al. 2023, Long et al. ms). As methods improve, organisations should use them to assess whether systems are likely to be conscious at several stages of development. As Shevlane et al. (2023) describe in the context of evaluations for AI safety, there are reasons to evaluate systems before and during training, before deployment, and later, after deployment, when more is known about their capabilities. The process of assessing systems for consciousness should be formally instituted in organisations' policies and should be audited by independent experts.

It is likely to be valuable for organisations to also consult with outside experts on decisions about whether to proceed with projects that may involve building conscious systems. Expert consultants can provide advice that can help to make the cost-benefit judgements discussed in the previous section, especially by providing alternative perspectives reflecting different biases and concerns from those inside the organisation. In making some momentous decisions organisations might also engage with authorities and the public (Birch, 2024).

In addition to assessing existing systems for consciousness and reflecting carefully on costs and benefits before developing new ones, organisations should seek to make gradual, limited progress in capacities linked to consciousness in developing new systems. The purpose is to minimise the risk of developing systems with the capacity for much richer conscious experiences than we realise. The potential problem to be avoided here relates to the concept of *overhangs*, which has been identified in AI safety research: underexplored systems may have hidden capabilities, or latent attributes that would allow leaps in performance if unlocked (Dafoe, 2018).

Putting these points together, we advocate a phased approach in which organisations work to understand the systems they have already built before moving on to new projects. They should then apply careful scrutiny to proposals, considering whether the proposed systems might be conscious and whether new features intended to enhance capabilities are warranted. We summarise this view in our third principle:

    3. Phased approach: Organisations should pursue a phased development approach, progressing gradually towards systems that are more likely to be conscious or are expected to undergo richer conscious experiences. Throughout this process, organisations should (i) implement strict and transparent risk and safety protocols and

(ii) consult with external experts to understand the implications of their progress and decide whether and how to proceed further.

**4.4 Knowledge Sharing: Transparency with Limits**

Given that the kind of AI consciousness research we would endorse has the objective of improving understanding, research organisations should share what they learn. Knowledge sharing makes for faster progress towards understanding through collaboration and scrutiny, and is essential for the value of understanding to be realised. In this case, researchers and authorities can only use knowledge to protect potential AI moral patients and promote public understanding if they can access it. As far as possible, research organisations should make information about their work available to the public, authorities and the research community.

However, there are limits to responsible knowledge sharing. If a research team succeeded in building a system that they believed was conscious, they should not generally make the full technical details of the system public, because this would make it possible for others who might mistreat the system to replicate it. This would be especially important if the system had capabilities that would incentivise its replication and (mis)use. In cases where information is sufficiently sensitive, for reasons of this kind, it should be protected and made available only to vetted experts and authorities.

There are some areas of research where the information hazards are so great that the research should not be conducted (Bostrom, 2011). For example, some forms of research on biological weapons are prohibited by the Biological Weapons Convention (United Nations, 1972). However, as we have argued in responding to Metzinger's moratorium proposal, we do not believe that AI consciousness is a case of this kind. Some research on AI consciousness may be beneficial enough to be worth conducting even though it would generate hazardous information, partly because this sensitive information could be adequately protected.

Our fourth principle is:

4. Knowledge sharing: Organisations should have a transparent knowledge sharing protocol that requires them to (i) make information available to the public, the research community and authorities, but only insofar as this is compatible with (ii) preventing irresponsible actors from acquiring information that could enable them to create and deploy conscious AI systems that might be mistreated or cause harm.

**4.5 Communication: Acknowledging Uncertainty**

How research organisations communicate about AI consciousness matters because of the dangers of poorly-informed public opinion on this topic. Organisations' first priority in communications should be to avoid misleading the public. There are various ways in which communications might be misleading.

One potential problem is overconfident dismissals of the possibility of AI consciousness. It may be tempting for research organisations to dismiss this possibility so as to avoid disruptive attention. As well as being misleading, however, overconfident dismissals may discourage research and voluntary and legal regulation. They suggest that AI consciousness is not a legitimate topic for researchers or policy-makers to attend to. Overconfident dismissals can come either in communications directly from organisations themselves, or in outputs from AI systems. For example, an LLM-powered chatbot might be made to insist that it could not possibly be conscious because it is an AI model. This would be misleading because, as we explained in section 2, many experts believe that it is possible for AI systems to be conscious. Preferable approaches to dealing with user queries on this topic could indicate uncertainty or refer to the best available evidence, perhaps by linking to an FAQ page.

Overconfident dismissals are an instance of a broader potential problem, which is failure to acknowledge the level of uncertainty in this area. Thus, for example, it is valuable for research organisations to state publicly why they take their current systems not to be conscious—assuming they do—but there is no benefit in making such statements in excessively confident terms. Similarly, it is helpful for organisations to be open about the theories that support their research (e.g. saying that they will seek to build AI systems equipped with global workspaces), but unhelpful for them to present these theories as known solutions to the problems of consciousness.

Organisations might attempt to attract attention and investment by promising to build a conscious system, but this would be both misleading, given the attendant uncertainty, and problematic in that it would present the creation of conscious AI systems as a prestigious scientific achievement. Perhaps this would be an achievement deserving significant prestige, but presenting it in this way encourages an irresponsible pursuit of consciousness itself. One can imagine a race between rival labs or countries to develop a conscious AI system. If understanding consciousness is expected to lead to building smarter, more efficient, or safer AI, then it would be reasonable for organizations to promote their work by highlighting consciousness. However, the phrase "Safe AI through understanding consciousness" does not frame AI consciousness as an exciting goal in itself, unlike a mission statement such as "Our mission is to solve consciousness," which directly emphasises that ambition.

Finally, while it's important for research organisations to recognise the potential harm of creating and mistreating AI moral patients, they must also remain mindful of other significant risks posed by AI. The focus on AI consciousness should not overshadow pressing concerns related to AI safety (Bostrom, 2014, Carlsmith, 2021, Alfonseca et al. 2021) and AI ethics (Zhang et al. 2021, Gabriel et al. 2024), For instance, the possibility of AI consciousness might become a distraction by diverting resources and attention away from other challenges, such as the development of robust AI safety measures, the mitigation of algorithmic biases, and the ethical implications of AI in decision-making processes.

Our recommendations for communication are summarised in our final principle:

5. Communication: Organisations should refrain from making overconfident or misleading statements regarding their ability to understand and create conscious AI. They should acknowledge the inherent uncertainties in their work, recognise the risk of mistreating AI moral patients, and be aware of the potential impact that communication about AI consciousness can have on public perception and policy making.

## 5. Maintaining Responsible Behaviour

Any organisation engaged in AI consciousness research will be made up of individuals with a variety of concerns and incentives, and these are liable to change over time. Even if such organisations initially endorse the five principles that we outline in this paper, they could be subject to incentives to abandon them in future. For example, a company may have a strong commercial incentive to ignore indications of consciousness in one of its systems if giving proper attention to these indications would interfere with deployment. Or an organisation may be tempted to pursue AI consciousness, and advertise themselves as doing so, if this will win them desirable attention. Furthermore, research always requires funding, so organisations will always be under pressure to please investors or other funders.

This means that it is important for organisations that currently aim to pursue AI consciousness research responsibly to act to ensure that they will continue to behave responsibly in the future. While it may be difficult for organisations to bind themselves irrevocably, it is possible to institute policies that will disincentivise irresponsible behaviour in the future.

Although different strategies are likely to be appropriate for different organisations, the following suggestions illustrate the kinds of actions we think organisations should consider.

First, one element of a strategy might be to make a public commitment to the principles outlined here, or to a similar set. This kind of outward-facing action could be supplemented by helping to establish or support external organisations whose function is to help promote responsible behaviour concerning AI consciousness. For example, a large company could provide funding for research elsewhere on the ethics of AI consciousness, or spin out a company specialising in consciousness evaluations.[9]

One function of such external organisations might be to communicate with the public, experts and authorities in ways that are more difficult for large research organisations themselves. However, external organisations could also audit AI consciousness research, as we have suggested. For either of these functions to be performed effectively, external organisations must be independent, as far as possible, from those they are auditing or potentially criticising, so support must be set up in a way that promotes this independence—for example, an auditor could provide services to several competing companies.

Second, research organisations should develop policies for reviewing projects and making choices that require attention to the ethical issues and principles discussed in this paper. They could also write conditions concerning responsibility into their codes of institutional values and practices, and into the rubrics they use to assess employee performance. Somewhat more substantively, they could appoint non-executive directors with the role of monitoring the organisation's adherence to principles like those described here, and using their power to maintain it.

Ultimately, we recognize that these measures may be ineffective in cases where strong incentives drive irresponsible behavior. However, their value lies in tipping the balance toward responsibility, increasing the likelihood that the ethical course of action will be taken more often. Nonetheless, some organisations engaged in AI consciousness research, or other forms of advanced AI development, may not be convinced of the importance of the principles we have outlined. As a result, legal interventions will likely become necessary, either to prohibit the creation or use of conscious AI systems or to provide them with legal protections against mistreatment.

---

[9] Comparable to safety evaluations providers such as METR.

## 6. Conclusion

If building conscious AI systems is becoming a realistic possibility, then organisations involved in advanced AI research should adopt policies addressing this prospect. Moreover, some organisations may justifiably wish to explore AI consciousness, but they must carefully consider how to do so responsibly. To guide this effort, we have proposed five principles for responsible research in this emerging era, in which conscious AI seems feasible but there is considerable uncertainty and risk.

**Acknowledgements**

Patrick Butlin's work on this research was funded by Conscium. [Further acknowledgements to come.]

**References**

Alfonseca, M., Cebrian, M., Anta, A. F., Coviello, L., Abeliuk, A., & Rahwan, I. (2021). Superintelligence cannot be contained: Lessons from computability theory. *Journal of Artificial Intelligence Research,* 70, 65-76.

Bach, T. A., Khan, A., Hallock, H., Beltrão, G., & Sousa, S. (2022). A Systematic Literature Review of User Trust in AI-Enabled Systems: An HCI Perspective. *International Journal of Human–Computer Interaction*, 40(5), 1251–1266.

Bajgar, O., & Horenovsky, J. (2023). Negative human rights as a basis for long-term AI safety and regulation. *Journal of Artificial Intelligence Research*, 76, 1043-1075.

Bayne, T., Seth, A. K., Massimini, M., Shepherd, J., Cleeremans, A., Fleming, S. M., ... & Mudrik, L. (2024). Tests for consciousness in humans and beyond. *Trends in cognitive sciences*.

Bengio, Y. (2017). The consciousness prior. *arXiv preprint arXiv:1709.08568*.

Bentham, J. (1789). *An Introduction to the Principles of Morals and Legislation*.

Birch, J. (2024). *The Edge of Sentience*. Oxford University Press.

Bostrom, N. (2011). Information hazards: A typology of potential harms from knowledge. *Review of Contemporary Philosophy*, (10), 44-79.

Bostrom, N. (2014). *Superintelligence: Paths, Dangers, Strategies*. Oxford University Press.

Bostrom, N., & Shulman, C. (2022). Propositions concerning digital minds and society. https://nickbostrom.com/propositions.pdf

Bradford, G. (2023). Consciousness and welfare subjectivity. *Noûs*, 57(4), 905-921.

Bradley, A., & Saad, B. (ms). AI alignment and AI ethical treatment: Ten challenges. https://docs.google.com/document/d/1o-TdNsgcdY57ev5T_Nabu6Ipf0PpCE9Ac7NjJEHw5YI/edit?usp=sharing

Butlin, P., Long, R., Elmoznino, E., Bengio, Y., Birch, J., Constant, A., ... & VanRullen, R. (2023). Consciousness in artificial intelligence: insights from the science of consciousness. *arXiv preprint arXiv:2308.08708*.

Cao, R. (2022). Multiple realizability and the spirit of functionalism. *Synthese*, 200, 506.

Carlsmith, J. (2022). Is power-seeking AI an existential risk?. *arXiv preprint arXiv:2206.13353*.

Carruthers, P. (2018). Valence and value. *Philosophy and Phenomenological Research*, 97(3), 658-680.

Chalmers, D. J. (2022). *Reality+: Virtual Worlds and the Problems of Philosophy*. Penguin UK.

Chalmers, D. (2023). Could a large language model be conscious? *Boston Review*. https://www.bostonreview.net/articles/could-a-large-language-model-be-conscious/

Colombatto, C., & Fleming, S. M. (2024). Folk psychological attributions of consciousness to large language models. *Neuroscience of Consciousness*, 2024(1), niae013.

Dafoe, A. (2018). AI governance: A research agenda. https://cdn.governance.ai/GovAI-Research-Agenda.pdf

Dehaene, S., Lau, H., & Kouider, S. (2017). What is consciousness, and could machines have it?. *Science*, 358(6362), 486-492.

Dossa, R. F. J., Arulkumaran, K., Juliani, A., Sasai, S., & Kanai, R. (2024). Design and evaluation of a global workspace agent embodied in a realistic multimodal environment. *Frontiers in Computational Neuroscience*, 18, 1352685.

Elamrani, A., & Yampolskiy, R. V. (2019). Reviewing tests for machine consciousness. *Journal of Consciousness Studies*, 26(5-6), 35-64.

Franklin, S., & Graesser, A. (1999). A software agent model of consciousness. *Consciousness and Cognition*, 8(3), 285-301.

Friston, K. (2013). Life as we know it. *Journal of the Royal Society Interface* 10: 20130475.

Gabriel, I., Manzini, A., Keeling, G., Hendricks, L. A., Rieser, V., Iqbal, H., ... & Manyika, J. (2024). The ethics of advanced AI assistants. *arXiv preprint arXiv:2404.16244*.

Godfrey-Smith, P. (2016). Mind, matter and metabolism. *Journal of Philosophy* 113, 481-506.

Godfrey-Smith, P. (2023). Nervous systems, functionalism and animal minds. https://petergodfreysmith.com/wp-content/uploads/2023/12/NYU-Oct-2023-Animals-AI-Functionalism-paper-Post-C3.pdf

Graziano, M. S. (2017). The attention schema theory: A foundation for engineering artificial consciousness. *Frontiers in Robotics and AI*, 4, 60.

Jaegle, A., Gimeno, F., Brock, A., Vinyals, O., Zisserman, A., & Carreira, J. (2021a). Perceiver: General perception with iterative attention. In *International conference on machine learning* (pp. 4651-4664). PMLR.

Jaegle, A., Borgeaud, S., Alayrac, J. B., Doersch, C., Ionescu, C., Ding, D., ... & Carreira, J. (2021b). Perceiver IO: A general architecture for structured inputs & outputs. *arXiv preprint arXiv:2107.14795*.

Juliani, A., Kanai, R., & Sasai, S. S. (2022). The perceiver architecture is a functional global workspace. *Proceedings of the Annual Meeting of the Cognitive Science Society* (Vol. 44, No. 44).

Kagan, S. (2019). *How to Count Animals, More or Less*. Oxford University Press.

Kiškis, M. (2023). Legal framework for the coexistence of humans and conscious AI. *Frontiers in Artificial Intelligence*, 6.

Koch, C. (2023). What does it 'feel' like to be a chatbot? *Scientific American*. https://www.scientificamerican.com/article/what-does-it-feel-like-to-be-a-chatbot/

Lau, H. (2022). *In Consciousness We Trust: The Cognitive Neuroscience of Subjective Experience*. Oxford University Press.

LeDoux, J., Birch, J., Andrews, K., Clayton, N. S., Daw, N. D., Frith, C., ... & Vandekerckhove, M. M. (2023). Consciousness beyond the human case. *Current Biology*, 33(16), R832-R840.

Liao, B., Pardo, P., Slavkovik, M., & van der Torre, L. (2023). The jiminy advisor: Moral agreements among stakeholders based on norms and argumentation. Journal of Artificial Intelligence Research, 77, 737-792.

Long, R. (2023). Introspective capabilities in large language models. *Journal of Consciousness Studies*, 30(9-10), 143-153.

Long, R., Sebo, J., Sims, T., Butlin, P., Harding, J., & Pfau, J. (ms). Towards assessments for AI moral patienthood.

Liu, D., Bolotta, S., Zhu, H., Bengio, Y., & Dumas, G. (2023). Attention schema in neural agents. *arXiv preprint arXiv:2305.17375*.

Metzinger, T. (2021). Artificial suffering: An argument for a global moratorium on synthetic phenomenology. *Journal of Artificial Intelligence and Consciousness*, 8(01), 43-66.

Michel, M., & Lau, H. (2021). Higher-order theories do just fine. *Cognitive Neuroscience*, 12(2), 77-78.

Papineau, D. (2023). The coming moral crisis and what to do about it. https://www.davidpapineau.co.uk/unpublished-talks.html

Petersen, S. (2007). The ethics of robot servitude. *Journal of Experimental & Theoretical Artificial Intelligence*, 19(1), 43-54.

Petersen, S. (2011). Designing people to serve. In *Robot Ethics: The Ethical and Social Implications of Robotics*, 283-298. MIT Press.

Resnik, D. B. (2018). *The Ethics of Research with Human Subjects: Protecting People, Advancing Science, Promoting Trust*. Springer.

Roelofs, L. (2023). Sentientism, motivation, and philosophical Vulcans. *Pacific Philosophical Quarterly*, 104(2), 301-323.

Russell, W. M. S., Burch, R. L., & Hume, C. W. (1959). *The Principles of Humane Experimental Technique*. Methuen.

Schneider, S. (2019). *Artificial You: AI and the Future of Your Mind*. Princeton University Press.

Schwitzgebel, E. (2023). The coming robot rights catastrophe. *Blog of the APA*. https://blog.apaonline.org/2023/01/12/the-coming-robot-rights-catastrophe/

Schwitzgebel, E., & Garza, M. (2020). Designing AI with rights, consciousness, self-respect and freedom. In *Ethics of Artificial Intelligence*, 459-479. Oxford University Press.

Sebo, J., & Long, R. (2023). Moral consideration for AI systems by 2030. *AI and Ethics*, 1-16.

Seth, A. (2021). *Being You: A New Science of Consciousness*. Penguin.

Seth, A. (2024). Conscious artificial intelligence and biological naturalism. https://doi.org/10.31234/osf.io/tz6an

Shanahan, M. (2006). A cognitive architecture that combines internal simulation with a global workspace. *Consciousness and Cognition*, 15(2), 433-449.

Shanahan, M. (2024). Simulacra as conscious exotica. *arXiv preprint arXiv:2402.12422*.

Shepherd, J. (2018). *Consciousness and Moral Status*. Taylor & Francis.

Shevlane, T., Farquhar, S., Garfinkel, B., Phuong, M., Whittlestone, J., Leung, J., ... & Dafoe, A. (2023). Model evaluation for extreme risks. *arXiv preprint arXiv:2305.15324*.

Shevlin, H. (ms.). All too human? Identifying and mitigating ethical risks of social AI. https://philarchive.org/rec/SHEATH-4

Shiller, D. (2024). Functionalism, integrity and digital consciousness. *Synthese* 203, 47.

Shulman, C., & Bostrom, N. (2021). Sharing the world with digital minds. In *Rethinking Moral Status*, 306-326. Oxford University Press.

Singer, P. (1990). *Animal Liberation (2nd Edition)*. HarperCollins.

Smithies, D. (forthcoming). Hedonic consciousness and moral status. In *Oxford Studies in Philosophy of Mind*.

Solms, M. (2021). *The Hidden Spring: A Journey to the Source of Consciousness*. Profile Books.

United Nations (1972). *Convention on the prohibition of the development, production, and stockpiling of bacteriological (biological) and toxin weapons and on their destruction*.

Verschure, P. F. (2016). Synthetic consciousness: the distributed adaptive control perspective. *Philosophical Transactions of the Royal Society B: Biological Sciences*, 371(1701), 20150448.

Zhang, B., Anderljung, M., Kahn, L., Dreksler, N., Horowitz, M. C., & Dafoe, A. (2021). Ethics and governance of artificial intelligence: Evidence from a survey of machine learning researchers. *Journal of Artificial Intelligence Research*, 71, 591-666.